\documentclass[a4paper,conference]{IEEEtran}

\usepackage{geometry}
\usepackage[utf8]{inputenc} 
\usepackage[T1]{fontenc}    
\usepackage{hyperref}       
\usepackage{url}            
\usepackage{booktabs}       
\usepackage{amsfonts}       
\usepackage{nicefrac}       
\usepackage{microtype}      

\usepackage{graphicx}
\graphicspath{{./}}
\usepackage{bm}
\usepackage{amsmath}
\usepackage{color}
\usepackage{colortbl}
\usepackage{multirow}
\usepackage{diagbox}

\definecolor{tableRowColor}{gray}{0.9}
\definecolor{ourColor}{gray}{0.97}
\definecolor{tableColor}{gray}{0.9}
\newcommand{\tabincell}[2]{\begin{tabular}{@{}#1@{}}#2\end{tabular}}
\newcommand{\etal}{et al. }

\title{Long-term Multi-granularity Deep Framework for Driver Drowsiness Detection}

\begin{document}
\author{\IEEEauthorblockN{Jie Lyu}
	\IEEEauthorblockA{Xi'an Jiaotong University\\
		Email: jiejielyu@outlook.com}
	\and
	\IEEEauthorblockN{Zejian Yuan}
	\IEEEauthorblockA{Xi'an Jiaotong University\\
		Email: yuan.ze.jian@xjtu.edu.cn}
	\and
	\IEEEauthorblockN{Dapeng Chen}
	\IEEEauthorblockA{Xi'an Jiaotong University\\
		Email: dapengchenxjtu@foxmail.com}
}
\maketitle

\begin{abstract}
   For real-world driver drowsiness detection from videos, the variation of head pose is so large that the existing methods on global face is not capable of extracting effective features, such as looking aside and lowering head. Temporal dependencies with variable length are also rarely considered by the previous approaches, e.g., yawning and speaking. In this paper, we propose a Long-term Multi-granularity Deep Framework to detect driver drowsiness in driving videos containing the frontal faces. The framework includes two key components: (1) Multi-granularity Convolutional Neural Network (MCNN), a novel network utilizes a group of parallel CNN extractors on well-aligned facial patches of different granularities, and extracts facial representations effectively for large variation of head pose, furthermore, it can flexibly fuse both detailed appearance clues of the main parts and local to global spatial constraints; (2) a deep Long Short Term Memory network is applied on facial representations to explore long-term relationships with variable length over sequential frames, which is capable to distinguish the states with temporal dependencies, such as blinking and closing eyes. Our approach achieves $90.05\%$ accuracy and about $37$ fps speed on the evaluation set of the public NTHU-DDD dataset, which is the state-of-the-art method on driver drowsiness detection. Moreover, we build a new dataset named FI-DDD, which is of higher precision of drowsy locations in temporal dimension.
\end{abstract}

\IEEEpeerreviewmaketitle

\section{Introduction}
It is reported that about 1.24 million people die on roads every year, while driver drowsiness accounts for 6\% \cite{WHO2013} of them.
Driver drowsiness indicates a driver is lack of sleep, which can be detected by the variation of physiological signal \cite{WangJJ2008}, vehicle trajectory \cite{ColicA2014,Rezaei2014} and facial expressions \cite{NakamuraMM13}. However the first two methods are hard to satisfy the requirement of convenience and timeliness. Drowsiness can be reflected by facial expression, such as nodding, yawning and closing eyes. We therefore aim to develop a drowsiness detection method based on video. Video-based method is possible to give the warning prompts and receive the driver's feedback in time, being of great value in practice.

Video-based drowsiness detection is still full of challenges, mainly stemming from the illumination condition change, head pose variation, and temporal dependencies. 
In particular, the large variation of head pose causes serious deformation of facial shape, which makes it difficult to extract effective spatial representations.
Conventionally, approach based on aligned facial points \cite{NakamuraMM13} is a better way to represent drowsy features, however, ignoring temporal relationships means it cannot distinguish blinking and closing eyes.
Spatio-temporal descriptor \cite{AkroutM15} is proposed to collect spatial and temporal features but not good at distinguishing states with long-term dependencies, such as yawning and speaking. 
Besides, these handcrafted descriptors are not enough powerful to describe large variation of head pose and classify confusing states, e.g., looking aside and lowering head lead to large pose variation, while yawning and laughing are similar but belong to different states.
Recently, deep learning methods are widely used to learn facial spatial representations automatically from global face \cite{YuPLJ16, HuynhPK16, ShihH16}.
Nevertheless, the global face without well alignment is weak to provide effective representations for large pose variation. Moreover, it is not flexible to fuse the configurations of local regions and concentrate representations on the most important parts such as eyes, nose and mouth on which the majority of drowsy information focuses. 
It is another challenge to distinguish easy-to-confuse states, such as blinking and closing eyes. 
3D-CNN with fixed time windows \cite{YuPLJ16} tried to describe spatial and temporal features, but it does not have enough capability to model long-term relationships with variable time length.

We propose a Long-term Multi-granularity Deep Framework (LMDF) to detect driver drowsiness from well-aligned facial patches. 
Our method applies alignment technology to obtain the well-aligned facial patches over frames, and these patches mainly locate in the informative regions that supply critical drowsy information.
A group of parallel convolutional paths are applied on the patches, and the outputs of these layers are fused by a fully connected layer to generate spatial representations, which is named as Multi-granularity Convolutional Neural Network (MCNN).
MCNN is able to fuse appearance of those well-aligned patches and capture local to global constraints. 
To explore temporal dynamical characteristics, a deep Long Short Term Memory (LSTM) network is applied to the spatial representations over sequential frames, which can distinguish the states with temporal relationships, such as yawning and laughing, blinking and closing eyes. 
The proposed method can thus not only extract effective facial representations on single-frame images, but also mine temporal clues on videos.

The contributions of our approach are mainly in three aspects:
(1) We propose a Long-term Multi-granularity Deep Framework to learn facial spatial features and their long-term temporal dependencies. 
(2) We propose MCNN to learn the facial representations from the most important parts, which makes the detector robust to large pose variation. 
(3) We build a Forward Instant Driver Drowsiness Detection (FI-DDD) dataset with higher precision of drowsy locations in temporal dimension, which is a good test bed for evaluating practical systems that are required to detect drowsiness in time.

\section{Related Work}
Driver drowsiness detection is becoming a hot topic of Advanced Driver Assistant System (ADAS). 
Many traditional methods are applied to deal with this problem.
The change of pupil diameter was utilized by Shirakata \etal \cite{TKJY2010} to detect imperceptible drowsiness, which is effective but it is not convenient for a driver to take the equipment.
Nakamura \etal \cite{NakamuraMM13} utilized face alignment to estimate the degree of drowsiness via k-NN, which cannot achieve online performance.
Spatial-temporal features for driver drowsiness detection was proposed by Mahdi \etal \cite{AkroutM15}, which was based on hough transformation, cannot work well in practical driving environment.  
Besides, the representations of those methods are hand-crafted, which may be not flexible to adapt to complex situations faced in driving, while our method automatically learns facial representations, which is more effective to the practical task. 

Deep learning approaches such as CNN have achieved success in representing information on images \cite{KrizhevskySH12,SunWT14,LeCunYG2015}, and many researchers also applied CNN on driver drowsiness detection.
Park \etal \cite{ParkPKY16} combined the results of three existing networks by SVM to present the categories of videos, which cannot detect driver drowsiness online. 
3D-CNN is applied to extract spatial and temporal information by Yu \etal \cite{YuPLJ16}, and the methods can only capture features with fixed temporal window. 
The above two methods utilize global face image, which cannot flexibly configure those patches containing the majority of drowsy information. Moreover, they are hard to capture dependencies with variable temporal length.

Due to the well performance of LSTMs on sequential data \cite{MaoXYWY14a,AlexGAMG2013,ChernodubN16}, more and more researchers propose combinations of CNN and LSTMs to learn spatial and temporal representations of sequential frames. 
It is interesting that Liang M. \etal \cite{LiangH15} came up with convolutional layers with intra-layer recurrent connections to integrate the context information for object recognition. 
Jeff D. \etal \cite{DonahueHGRVDS15} provided a method which extracts visual features from images by CNN and learns the long-term dependencies from sequential data by LSTMs. 
Especially, the approach of Jiang W. \etal \cite{WangYMHHX16} processes image with CNN and models sequential labels by LSTMs concurrently, and then combines the two representations via projection layers. 
However, none of the above methods apply multi-granularity method to concentrate representations on important parts and flexibly fuse configurations of different regions.

Recently, Multi-granularity methods have achieved several excellent results in other applications of computer vision. 
Qing Li \etal \cite{LiQYMRL16} proposed temporal multi-granularity approach of action recognition. 
Their method achieved the state-of-the-art performance on action benchmarks, but cannot capture  detailed appearance clues and local to global spatial information. 
Dong C. \etal \cite{ChenCWS2013} applied multi-scale patches based on face alignment on face recognition. 
Dequan W. \etal \cite{WangSSZXZ15} utilized multi-granularity regions, detected by three granularities convolutional neural network, to generate multi-granularity descriptor for fine-grained categorization, but this method cannot process sequential frames. 
Different from the above, our method can capture spatial multi-granularity information and long-term temporal dependencies. 
Particularly, our MCNN can learn representations on the most significant regions from well-aligned multi-granularity patches, and the proposed method has achieved the state-of-the-art accuracy on NTHU-DDD dataset for driver drowsiness detection. 

\section{Our Approach}
\begin{figure*}[htbp]
	\centering
	\includegraphics[width=1.0\linewidth]{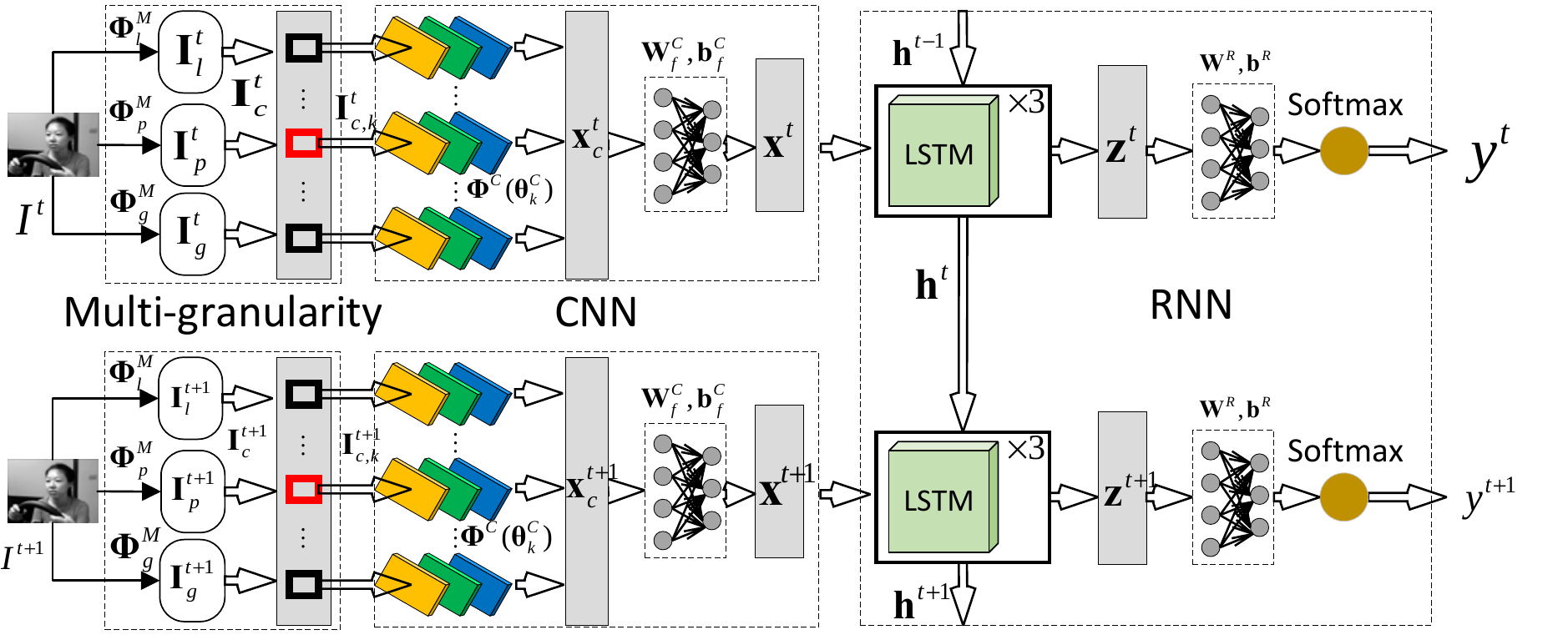}
	\caption{The long-term multi-granularity deep framework for driver drowsiness detection. The first stage is well-aligned multi-granularity patches which consist of local regions, main parts and global face. Parallel convolutional layers are well-applied to process patches respectively and fully connected layer fuses local and global clues and generates a representation, which is the second stage of the framework. The first two stages construct the Multi-granularity CNN (MCNN). Recurrent Neural Network (RNN) with multiple LSTM blocks mining the clues in temporal dimension together with a fully connected layer form the third stage.}
	\label{fig_framework}
\end{figure*}

The proposed method utilizes Multi-granularity Convolutional Neural Network (MCNN) to learn facial representations from single-frame images. 
The repreentations, extracted from well-aligned facial patches, contains both detailed appearance information of the main parts and local to global constraints. 
Furthermore, our approach takes advantages of a deep Long Short Term Memory (LSTM) network to explore dynamical characteristics of the facial representations from sequential frames. 
The detailed structure of our Long-term Multi-granularity Deep Framework combining MCNN and LSTMs is shown as Fig.\ref{fig_framework}.

\subsection{Well-aligned Multi-granularity Patches}
It is well known that drowsy information is focused on several main facial parts such as eyes, nose and mouth. 
Alignment provides an excellent way to extract well-aligned features over frames, which effectively represent facial drowsy states. 
Besides, global patch provides rough information to estimate the states of a driver's head and full face, which assists the decision of driver's drowsy states when the locations of parts are not precise. 
Our method takes advantages of local regions and global face at the same time.

We utilize face alignment technology to locate facial shape points.
Given an image $I^t$ with a face in the $t$-th frame, we detect landmark points of facial shape $S^t$ via regressing local binary features proposed by Ren \etal \cite{RenSQ2014}. 
From those points, it is convenient to get the locations of main parts and important local regions. 
According to center points and specific sizes of all regions, we crop those patches from the original image and resize them into the same size, which are the well-aligned multi-granularity patches as the input layers of the convolutional neural network. 

\begin{figure}[htbp]
	\centering
	\includegraphics[width=0.9\linewidth]{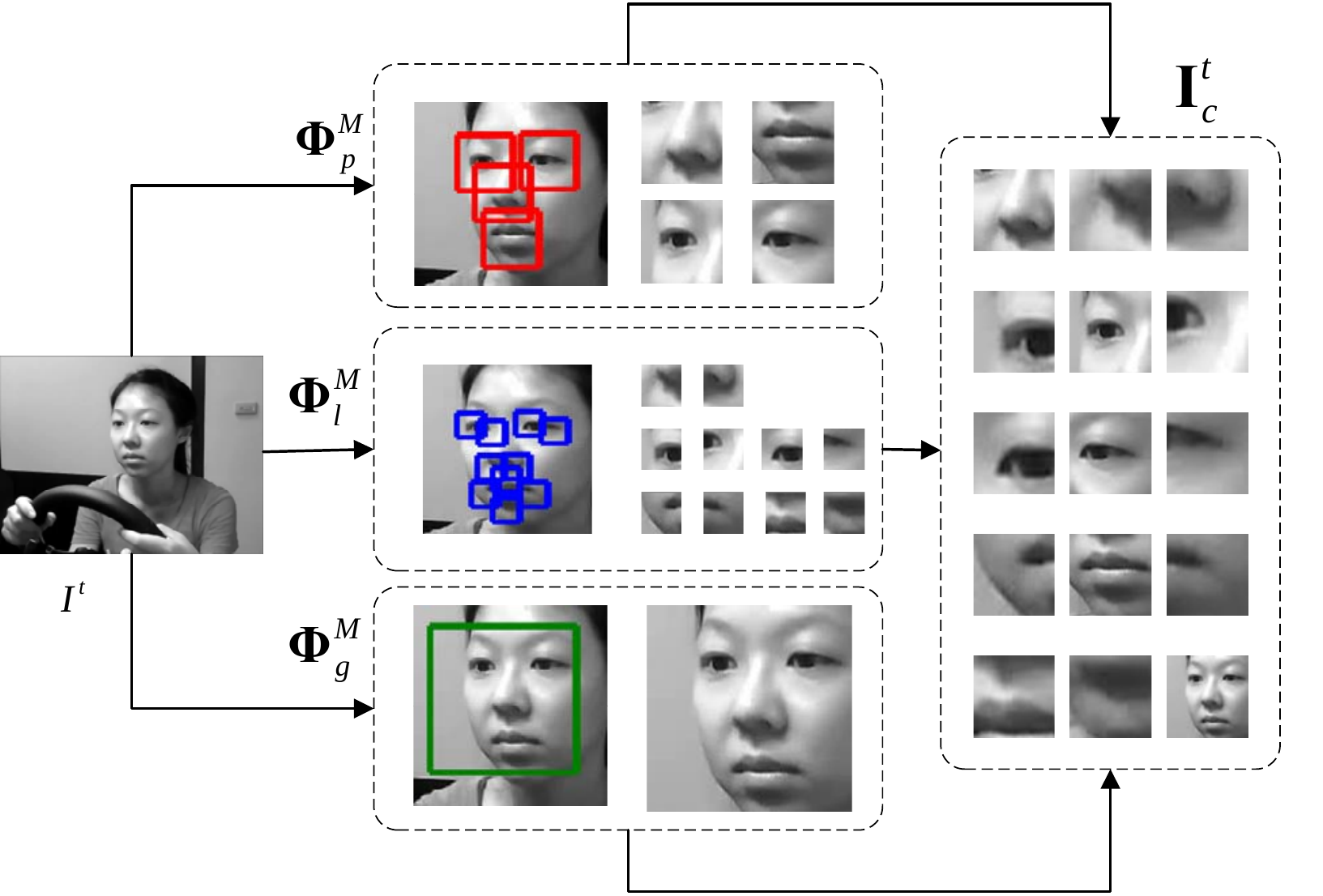}
	\caption{The procedure of extracting multi-granularity facial patches which include the main parts, local regions and global face such three granularities.}
	\label{fig_patch_extract}
\end{figure}

Those patches, including local regions, main parts, and the global face, are produced by three different mappings. Shown as Fig.\ref{fig_patch_extract}, a mapping ${\bf{\Phi}}^M_p$ can select center points of eyes, nose, and mouth from facial shape $S^t$, and crop patches of those parts from the input image $I^t$ with given sizes $s_p$. And the mapping still needs to convert the patches into an unified size $s_u$. Thus the single-granularity patches of those main parts ${\bf{I}}_p^t$ are generated. The operations of mapping ${\bf{\Phi}}^M_l$ and ${\bf{\Phi}}^M_g$ are similar to the mapping ${\bf{\Phi}}^M_p$, while the differences lie in the locations and sizes of regions. The mapping ${\bf{\Phi}}^M_l$ selects the corners of the eyes and mouth and the sides of the nose as the interest of regions with size $s_l$ and output local patches ${\bf{I}}_l^t$. A global facial region with size $s_g$ is chosen by the mapping ${\bf{\Phi}}^M_g$ which finally produces a global facial patch ${\bf{I}}_g^t$. Formally, the mappings are represented as
\begin{equation}
{\bf{I}}_i^t = {\bf{\Phi}}^M_i(I^t,S^t,s_i,s_u), i \in \{l,g,p\}.
\label{equ_patch_mapping}
\end{equation}

Processing the input image $I^t$ by the three mappings, we can obtain a set of well-aligned patches ${\bf{I}}_c^t$ consisting of the main parts, local regions and global face, and it is presented as 
\begin{equation}
{\bf{I}}_c^t = \{{\bf{I}}_{l,:}^t,{\bf{I}}_{p,:}^t,{\bf{I}}_{g,:}^t\},
\label{equ_patch_set}
\end{equation}
where ${\bf{I}}^t_{i,:},i\in\{l,p,g\}$ represents all elements of a patch set ${\bf{I}}^t_i$. 

Compared with the original image, the patches set ${\bf{I}}_c^t$, including both detailed appearance clues of parts and rough information of full face, have more advantages to describe the facial states. Meanwhile, the relations between local and global regions are implied, which is the basis of mining useful features. Therefore, we take the set of patches ${\bf{I}}_c^t$ as the input layer of CNN to learn effective representations. 

\subsection{Learning Facial Representations}
Our approach learns representations by convolutional neural network but not hand-crafted for its well performance in learning spatial features. We apply several convolutional layers to processing each one in the set of patches ${\bf{I}}_c^t$ independently. To fuse the information of all patches, a fully connected layer is arranged after all convolutional operations, which generates $N$-dimensional descriptors combining local and global clues.  

Every patch needs to be processed by convolutional operations at first. For a patch $I_{c,k}^t$, the $k$-th one of patch set ${\bf{I}}_c^t$ with length $L$, three convolutional layers are utilized to capture the spatial feature. The first one is made with convolution and rectified linear units (ReLU) activation followed by a max-pooling operation, which projects a normalized 3-channel image to a higher dimensional representation. Only convolution and ReLU activation are selected in the second layer to enlarge the dimension of representation sequentially. And the structure of the third convolutional layer is similar to the first layer but with different parameters to decrease the dimension. A representation ${\bf{x}}_k^t$ of the patch ${\bf{I}}_{c,k}^t$ can be generated by a mapping ${\bf{\Phi}}^C$ consisting of those convolutional layers with parameters ${\bm{\theta}}^C_k$, which is presented as
\begin{equation}
{\bf{x}}_k^t = {\bf{\Phi}}^C({\bm{\theta}}^C_k,{\bf{I}}_{c,k}^t),k=1,2,\dots,L,
\label{equ_convolution}
\end{equation}
where ${\bm{\theta}}^C_k$ is the $k$-th element of convolutional parameter set ${\bm{\theta}}^C$.

A fully connected layer is utilized to combine those representations extracted by the mapping ${\bf{\Phi}}^C$ from the set of patches. But before combining operation, we concatenate those representations into a long vector ${\bf{x}}_c^t$, formed as
\begin{equation}
{\bf{x}}_c^t = [{\bf{x}}_{k}^t], k=1,2,\dots,L.
\label{equ_cancat}
\end{equation} 

With a specific weighted matrix ${\bf{W}}^C_f$ and bias vector ${\bf{b}}^C_f$, the combining $N$-dimensional representation ${\bf{x}}^t$ can be presented by the fully connected layer as
\begin{equation}
{\bf{x}}^t = \max\{{\bf{W}}^C_f{{\bf{x}}_c^t}+{\bf{b}}^C_f,{\bf{0}}\},
\label{equ_fullconnect}
\end{equation}  
in which ${\bf{0}}$ is a zero vector.

The descriptor ${\bf{x}}^t$ contains not only detailed appearance information implied in every part, but also the constrained relations between local regions and global face. The effectiveness of the descriptor can be improved by appropriate objective functions and proper training methods.

Driver drowsiness detection is a binary classified problem, thus the state of an input frame is just drowsy or not. We label drowsiness with $1$ as the positive sample and normal state with $0$ as the negative sample. And a label $c$ are expressed with a one-hot vector ${\bf{y}}_c$, such as a vector $[0,1]$ means the positive label.

To train the parameters of the convolutional neural network, we project the representation ${\bf{x}}^t$ into the probabilities of each category $c\in\{0,1\},$ by another fully connected layer with weights ${\bf{W}}^C_p$ and a bias vector ${\bf{b}}^C_p$, and the probability vector ${\bf{p}}(c|{\bf{x}}^t,{\bf{W}}^C_p,{\bf{b}}^C_p)$ are normalized via a softmax layer.
The cross entropy which can indicate the correct rate of classification is selected as the objective function, and we utilize the adam optimizer to train the whole convolutional neural network. The visual representations can also be generated by the convolutional layers and the first fully connected layer.

\subsection{Exploring Dynamical Characteristics}
The representation ${\bf{x}}^t$ is extracted in a frame, while whether a driver is drowsy or not is judged by a certain period. 
We apply LSTMs to model the temporal dynamical characteristics of spatial representations on driver drowsiness detection.

A LSTM block consists of an input gate, a forget gate, an output gate and a memory cell. Because of the three gates, the LSTM block can learn long-term dependencies in sequential data and its parameters are easier to be trained. The memory cell can store long-term information in its vector, which can be rewritten or done other operations for the next time step. Besides, the number of hidden units should be chosen according to the dimension of the input representation ${\bf{x}}^t$.

We employ multiple layers LSTMs to mine the temporal features for driver drowsiness. A mapping ${\bf{\Phi}}^R$ containing three layers LSTMs with parameters ${\bm{\theta}}^R$ is utilized to explore temporal clues of the representation ${\bf{x}}^t$ generated by MCNN extractor and presents the hidden states ${\bf{h}}^t_3$ of the third layer as a representation containing temporal dependencies,  which is presented as
\begin{equation}
{\bf{h}}^t_3 = {\bf{\Phi}^R}({\bm{\theta}}^R,{\bf{h}}^{t-1},{\bf{x}}^t),
\label{equ_rnn_lstm}
\end{equation}
where ${\bf{h}}^{t-1}=\{{\bf{h}}^{t-1}_1,{\bf{h}}^{t-1}_2,{\bf{h}}^{t-1}_3\}$ is the parameter set of these LSTM blocks in the last step.

A fully connected layer with weight ${\bf{W}}^R$ and a bias vector ${\bf{b}}^R$ is used to project the output of the mapping ${\bf{\Phi}}^R$ into a two-dimensional vector that is then decoded by softmax operation to the probabilities ${\bf{p}}(c|{\bf{h}}^t_3,{\bf{W}}^R,{\bf{b}}^R)$ of the two categories.
To solve the parameters, we take advantage of Adam optimizer to train the LSTMs with cross-entropy objective function.

The label $y^t$ of the current frame can be predicted as the class with the maximum probability.
\begin{equation}
y^t=\arg \max_c {\bm{p}}(c|{\bf{h}}^t_3,{\bf{W}}^R,{\bf{b}}^R), c\in\{0,1\}
\label{equ_rnn_pre}
\end{equation}
Similarly, the labels ${\bf{y}}$ of the sequential data can be obtained. 

\section{Experiments}
A dataset named National TsingHua University Drowsy Driver Detection (NTHU-DDD) is provided on the challenge of ACCV2016 workshop for driver drowsiness detection, on which we compare our approach with others. To make the sequential labels close to the practical driving environments, we relabel the video set with instant detecting principle. A new dataset is generated from the relabeled video set and it is called Forward Instant Driver Drowsiness Detection (FI-DDD) on which we learn parameters and analyze the performance of several subnetworks. While the performance of our entire approach is evaluated on the original NTHU-DDD dataset, we thus train a set of parameters to achieve long-term memory performance. Finally, the accuracy $90.05\%$ is obtained by our Long-term Multi-granularity Deep Framework (LMDF) on the evaluation set of the NTHU-DDD dataset, and the proposed method achieves about 37 fps on GPU Tesla M40. 

\subsection{Dataset}
{\bf{NTHU-DDD Dataset:}}
The NTHU dataset includes five scenarios listed as glasses, no glasses, glasses at night, no glasses at night and sunglasses. The training set involves 18 volunteers consisting of 10 men and 8 women who act as drivers with four different states in every scenario, while the evaluation set has four volunteers including two men and two women. Non-sleepy videos contain only normal state, while sleepy videos combine normal and drowsy states together. Besides, blinking with nodding and yawning videos only record drowsy eyes and mouth respectively. NTHU-DDD dataset offers four annotation files recording the states of drowsiness, eyes, head and mouth for every video. Table \ref{tab_nthu_ddd_states} gives the labels of drowsiness and three main parts.
\begin{table}[htbp]
	\centering
	\caption{The labeled states of each part on NTHU-DDD dataset}
	\begin{tabular}{c|ccc}
		\hline
		\rowcolor{tableRowColor}
		\diagbox{sub-\\class}{label} & 0 & 1 & 2\\
		\hline
		drowsiness & normal   & drowsy & - \\
		\rowcolor{tableRowColor}
		eyes & normal  &sleepy  & -  \\
		head & normal   & nodding   & looking aside \\
		\rowcolor{tableRowColor}
		mouth & normal   & yawning   & talking \& laughing \\
		\hline
	\end{tabular}
	\label{tab_nthu_ddd_states}
\end{table}
It is worth emphasizing that the labels on NTHU-DDD dataset are long-term memory, which means that the states of a frame may depend on the frames in the previous several seconds. 

{\bf{FI-DDD Dataset:}}
A problem comes due to the long-term memory in NTHU-DDD, which is that a driver would still receive the warning prompts even if he had revised his drowsy states to the normal for a few seconds. 
At the same time, those labels are unable to locate the drowsy states with high precision in temporal dimension. 
To solve these problems, we relabel those videos with {\bf{instant principle}}, which means the latency is limited within $0.5$ second namely $15$ frames for $30$ FPS videos. 
Those typical states, such as closing eyes, yawning and lowering head, are still considered as one of the evidences to judge whether a frame is drowsy. 
Those videos are cut into several clips which contain only the drowsy or the normal states alternatively according to our labels. 
To describe the transitional states between the normal and the drowsy, we reserve ten normal frames at the head and the tail of every clip with drowsiness. 
We name the relabeled dataset with Forward Instant Driver Drowsiness Detection (FI-DDD) which includes 14 drivers on trainset and 4 ones on testset. 
The trainset of FI-DDD at day time has $157$ clips and the testset has $92$ clips, while at night scenarios, the trainset has $126$ ones and the testset has $75$ clips with about $530$ frames on average. 

{\bf{Static image set:}}
To train the parameters of CNN and analyze the effects of several factors, we build a static image set by sampling lots of frames from the FI-DDD dataset. 
The samples on the image set are labeled with drowsiness or normality, and the labels can almost indicate the truth states of the corresponding images, even if a small amount of images are matched with wrong labels due to lack of temporal dependence. 
The static image set has $7498$ images in day time, and trainset includes $5239$ images and testset has $2259$ images. 
It has $2653$ images in night scenario, and trainset includes $1750$ images, testset has $903$ images.

\subsection{Implementation Details}
{\bf{Face Alignment}}: We apply face alignment technology to locate those facial shape points for all videos. Face detection and tracking are combined to increase detecting rate and provide more accurate positions for faces on videos. Face alignment algorithm is based on those face positions. The face detector is from OpenCV and the approach of face tracking is proposed by Danelljan \etal \cite{DanelljanHKF14}. We implement the method of Ren \etal \cite{RenSQ2014} and retrain the model, and preprocess all videos to obtain the $51$ landmark points for every frame. Those frames with no face will be recognized as the empty and filled with zero coordinates for landmark points. 

{\bf{Multi-granularity}}: We obtain Multi-granularity patches considering two factors: different positions and different sizes. We design to choose $15$ positions from facial shape points, which are divided into three granularities: 1 global face with size $s_g=(160\times160)$, 4 main parts with size $s_p=(64\times64)$ and 10 local regions with size $s_l=(32\times32)$. The specific locations of all patches are shown as Fig.\ref{fig_patch_extract}. Before sent to CNN, those patches are resized to size $s_u=(64\times64)$, normalized to [-0.5, 0.5], and are converted to $3$ channels to ensure that our framework can process RGB data.

{\bf{Dataset Usage}}: A static image set, required for training the CNN parameters, are sampled from the videos of FI-DDD with a specific frame interval. The results of CNN is directly related to multi-granularity patches and CNN parameters, we thus analyze the effects of those factors on the static image set. While all experiments for analyzing the effects of LSTMs parameters are carried on FI-DDD dataset. To compare with the previous methods, we evaluate the proposed method on the evaluation set of NTHU-DDD dataset.

\subsection{Experimental Analysis}
To further explain the effects of alignment, multi-granularity and CNN extractor, several groups of experiments are conducted on the {\bf{static image set}}. We also provide experiments on {\bf{FI-DDD}} dataset to verify the effectiveness of LSTMs for detection drowsiness on video.  

\subsubsection{{\bf{The Importance of Alignment}}}
It is essential to carry out experiments to explain the significance of alignment and the effects of locating precision. 

\begin{figure}[htbp]
	\centering
	\includegraphics[width=1.0\linewidth]{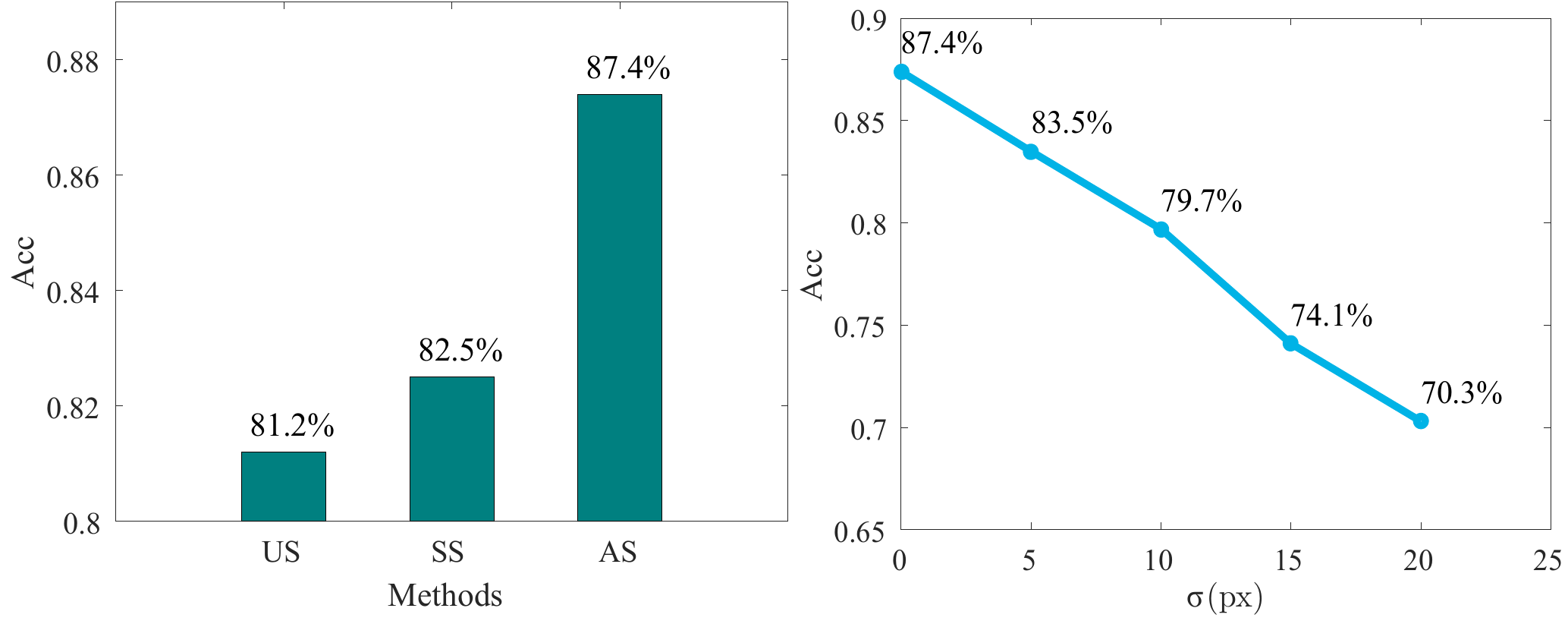}
	\caption{Left: the comparison of different sampling methods, sampling over uniform distribution({\bf{US}}), sampling specific locations({\bf{SS}}) and our proposed sampling with aligned positions({\bf{AS}}); Right:  the effect of alignment precision, $\sigma$ is the standard deviation of normal distribution. The results(Acc) are achieved by CNN on testset of the static image set.}
	\label{fig_fa_effect}
\end{figure}

{\bf{None-alignment $vs$ With Alignment:}}
We provide another two none-alignment methods to sample those multi-granularity patches in facial bounding box: Uniform Sampling ({\bf{US}}) and Specific Sampling ({\bf{SS}}). The corresponding sizes of our Aligned sampling ({\bf{AS}}) method and the two none-alignment ones are the same. Fig.\ref{fig_fa_effect}(Left) shows the comparison of {\bf{AS}}, {\bf{US}} and {\bf{SS}}. {\bf{AS}} considering alignment achieves the best accuracy $87.4\%$ on the testset of the static image set, which is $4.9\%$ higher than {\bf{SS}} method and $6.2\%$ higher than {\bf{US}} one. As a conclusion, alignment of facial patches, providing aligned representations, is an effective way to improve the accuracy on driver drowsiness detection.

{\bf{Effects of alignment precision:}}
We evaluate the effects of the alignment precision, and research the influence quantitatively by adding random noise with Gaussian distribution $N(0, \sigma)$ over the well-aligned facial points. Fig.\ref{fig_fa_effect}(Right) shows the results on testset of the static image set, from which, we discover that the accuracy is decreasing with the increasing standard deviation of noise and even less than $80\%$ if $\sigma\geq10$ px. While the accuracy is more than $83\%$ with $\sigma$ less than $5$ px, we make a conclusion that the proposed MCNN is robust to the corrupted locations if $\sigma\leq5$ px.

\subsubsection{{\bf{The Effects of Multi-granularity Patches}}}
Multi-granularity patches consist of local regions, main parts and global face. It is significant to conduct experiments and explain the importance of those granularities on driver drowsiness detection. We apply a fully connected layer and softmax operation to classify representations presented by MCNN extractor, and analyze the effects of multi-granularity patches by results of the classification.

\begin{figure}[htbp]
	\centering
	\includegraphics[width=0.85\linewidth]{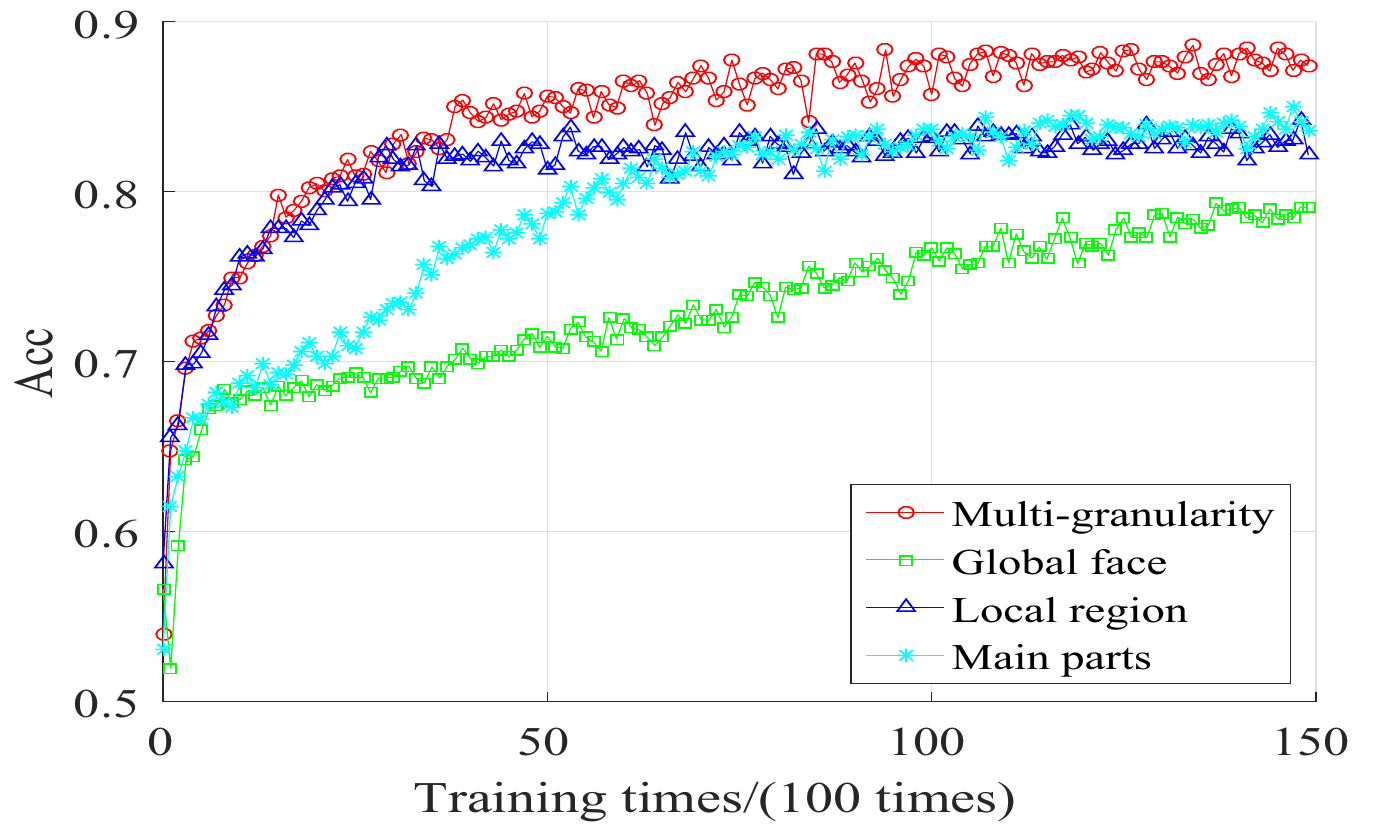}
	\caption{The comparison of different granularities, global face, main parts, local regions and multi-granularity patches. The Curve of accuracies(Acc) over training times are achieved by CNN with different granularities on testset of the static image set.}
	\label{fig_cnn_analysis}
\end{figure}

{\bf{Learning curve on different granularities:}}
We take four different granularities, listed as local regions, main parts, global face and the combination of the above, into account to analyze the effects of multi-granularity facial patches. Fig.\ref{fig_cnn_analysis} illustrates the comparisons of those granularities, from which, we know that the convergent speed of method with global face granularity is the slowest compared with the others, and that of local regions is the fastest. While multi-granularity method achieves good performance on both convergent speed and accuracy. Aligned points can achieve higher precision on those local regions with abundant boundary texture, which results in more aligned representations and easier being classified. Nevertheless, multi-granularity patches containing more aligned information is more effective on driver drowsiness detection. 

\begin{figure}[htbp]
	\centering
	\includegraphics[width=1.0\linewidth]{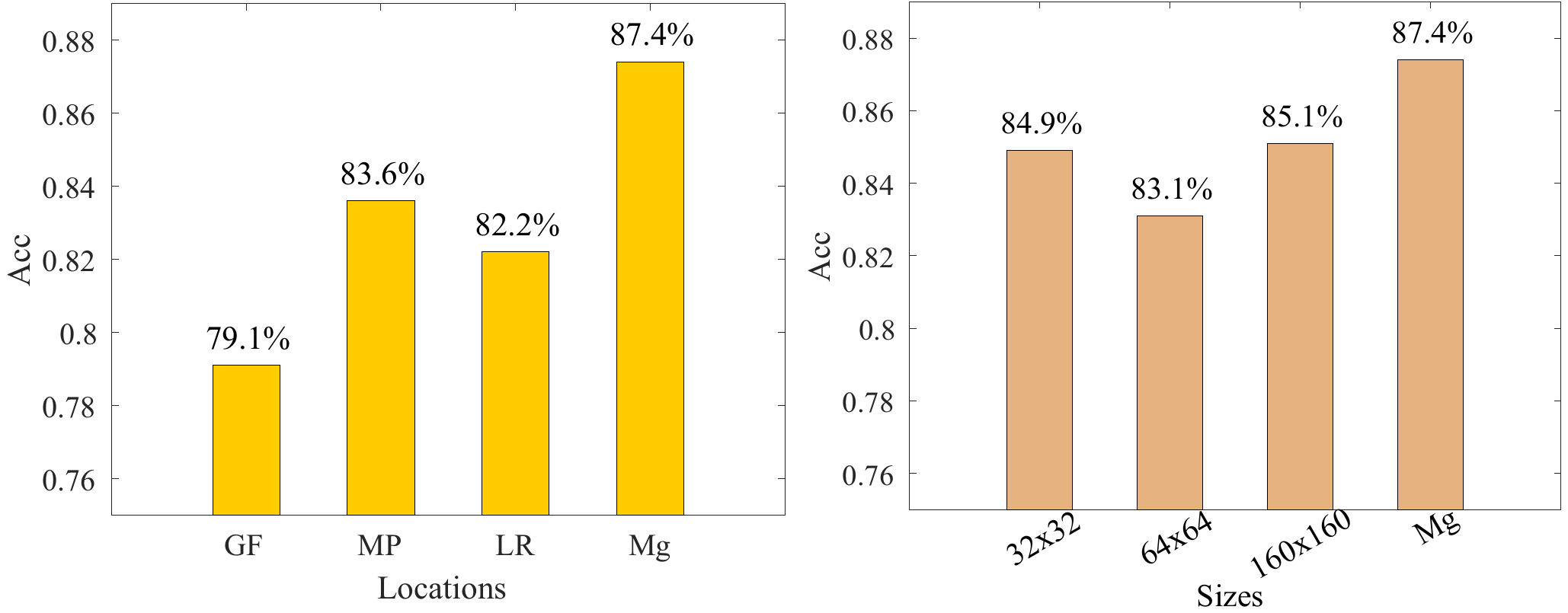}
	\caption{Left: The comparison of patches with different positions, GF-the global face, MP-main parts(eyes, nose and mouth), LR-local regions(the corner of eyes, the sides of nose and the boundary of mouth); Right: the comparison of patches with different sizes at all locations. Mg represents multi-granularity patches.}
	\label{fig_ls_effect}
\end{figure}

{\bf{Effects of positions and sizes:}}
We change the positions and sizes of facial patches respectively. Shown as Fig.\ref{fig_ls_effect}(Left), facial main parts, including eyes, nose and mouth, obtain the best accuracy $83.6\%$ compared with the other single-granularity method. Obviously, the combination of those three granularities achieves the best accuracy $87.4\%$. A conclusion comes that the most effective representation is extracted from the three main facial parts, while the fusion of local and global clues is an excellent way to obtain better facial representations. 

We set their sizes as the same and change the sizes to research the difference between single-size and multi-granularity methods, keeping the locations of these patches invariable. Fig.\ref{fig_ls_effect}(Right) shows different regions with different sizes achieve $2.3\%$ accuracy more than that of those single-size patches. The phenomenon is the result of that different physiological parts are of different sizes, e.g., the size of global face is bigger than single eye. The above analysis presents that multi-granularity method is an effective way to represent facial features.

\subsubsection{{\bf{The Parameters Selection of MCNN Extractor}}}
The structure parameters of the convolutional layers are listed as Table \ref{tab_conv_para}. A patch with size $64\times64$ processed by those convolutional layers is projected to a tensor with size $16\times16\times4$. And a representation of the patch is generated by reshaping the tensor to a $1024$-dimensional vector, which is the input of a fully connected layer.

\begin{table}[htbp]
	\centering
	\caption{The parameters of the three convolutional layers. Those structure parameters are the same in all parallel convolutional paths.}
	\begin{tabular}{c|cc}
		\hline
		\rowcolor{tableRowColor}
		Layers & Operations & Attributions\\
		\hline
		\multirow{3}{*}{1st} & Convolution & size:$[5\times5\times3\times32]$\\
		 & Activation & ReLU\\
		 & Max pooling & strides:$[2\times2]$\\
		\hline
		\multirow{3}{*}{2nd}& Convolution & size:$[5\times5\times32\times64]$\\
		 & Activation & ReLU\\
		 & Max pooling & Not Used\\
		\hline
		\multirow{3}{*}{3rd} & Convolution & size:$[5\times5\times64\times4]$\\
		 & Activation & ReLU\\
		 & Max pooling & strides:$[2\times2]$\\
		\hline
	\end{tabular}
	\label{tab_conv_para}
\end{table}
 
\begin{figure}[htbp]
	\centering
	\includegraphics[width=0.85\linewidth]{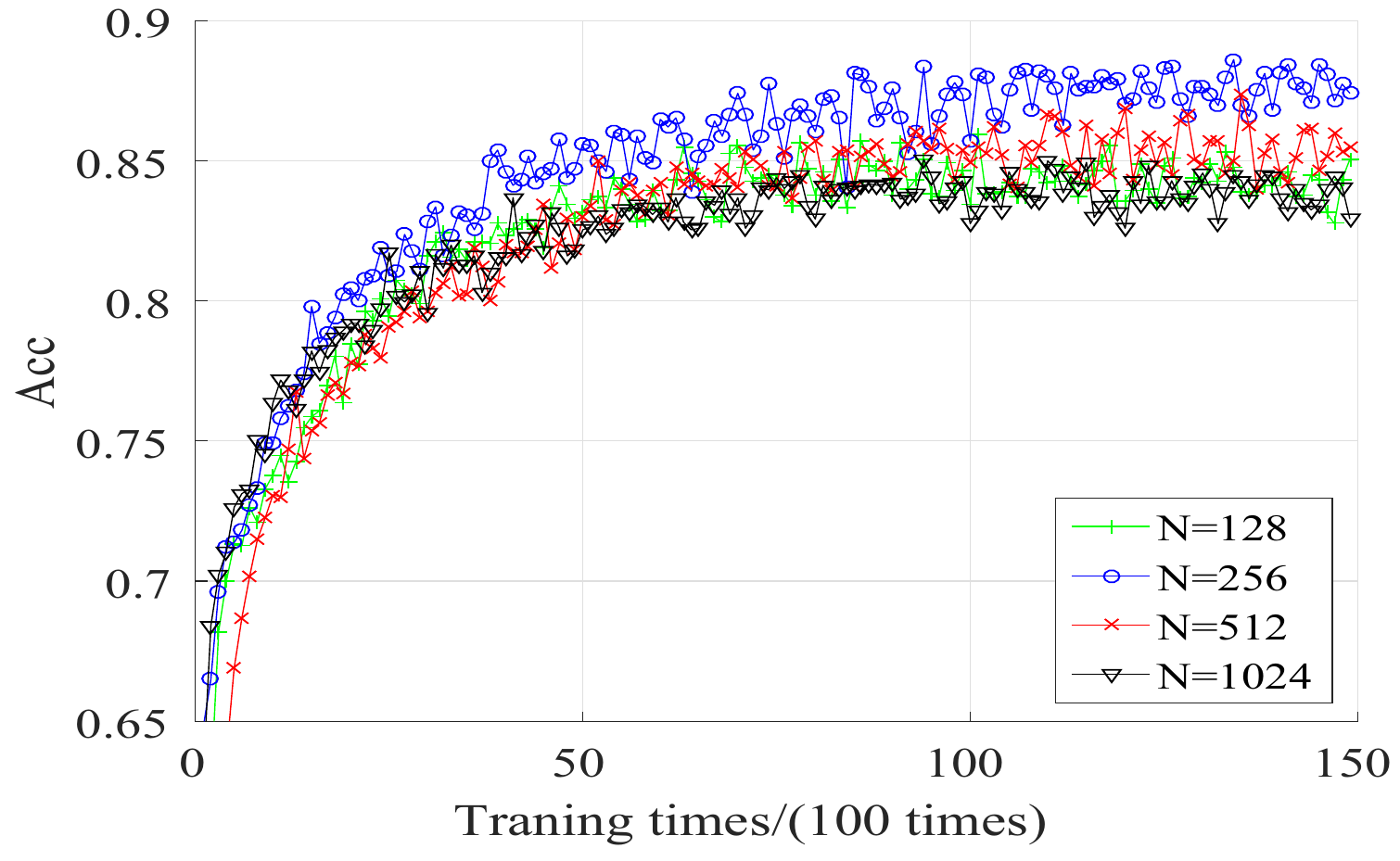}
	\caption{The comparison of different dimensional MCNN representations on accuracies and convergent performance achieved by CNN on testset of the static image set in daytime.}
	\label{fig_cnn_dim}
\end{figure}

A fully connected layer is applied to combine the multi-granularity clues and generate MCNN representations. The number of its hidden units $N$, namely the dimension of representation, has effects on the combination of those patches. Changing the number of hidden units $N$, we explore the relations between the dimension of MCNN representations and classification accuracy with well-aligned multi-granularity facial patches. The comparison of different dimensions is shown as Fig.\ref{fig_cnn_dim}, which indicates that the number of dimension almost has no influence on the convergent speed. But $256$-dimensional representations achieve the highest accuracy. Therefore it is reasonable for us choose the number of hidden units as $256$. 

\begin{table*}[htbp]
	\centering
	\caption{The comparison of different methods on the evaluation set of NTHU-DDD dataset with the detailed information of environments.}
	\begin{tabular}{cccccccc}
		\hline
		\rowcolor{tableRowColor}Methods & Platform 		 & Spatial Features 	        & Sequential Features 	& Speed 	& Accuracy\\
		Yu \etal \cite{YuPLJ16} 		& GPU 			 & \tabincell{c}{3D-DCNN} 		& feature fusion		& 24$\scriptsize{\sim}$32 fps 		& 72.60\% \\
		Park \etal \cite{ParkPKY16} 	& - 				& \tabincell{c}{DDD Network} 	& SVM					& - 		& 73.06\% \\
		MSTN \cite{ShihH16} 			& - 						& \tabincell{c}{CNN} 			& LSTMs					& 60 fps 	& 85.52\% \\
		\rowcolor{ourColor}Ours 		& \tabincell{c}{GPU-M40}  	& MCNN 							& LSTMs 				& 37 fps 	& 90.05\%\\
		\hline
	\end{tabular}
	\label{tab_performance}
\end{table*}

\subsubsection{{\bf{The Significance of LSTMs}}} 
We first apply MCNN to detect driver drowsiness on video, but it has no capacity to capture the temporal clues. MCNN+LSTMs is considered to deal with this drawback. It is necessary to compare the situation with LSTMs \cite{GreffSKSS15} and that without LSTMs for understanding the effects of LSTMs. All experiments at this part are carried on the FI-DDD dataset in day-time scenarios.

{\bf{Parameters setting:}}
The representations given by MCNN extractor are $256$-dimensional, and the number of hidden units in each LSTM block is equal to $256$.  The forget gate is enabled and the max memory step is set to $60$ frames. We randomly select a batch with $1000$ samples to train the LSTMs parameters with learning rate $3e^{-4}$. The fully connected layer projects the states of the last LSTM block to a $2$-dimensional vector which is decoded to the probability of drowsiness by a softmax operation.

\begin{figure}[htb]
	\centering
	\includegraphics[width=0.9\linewidth]{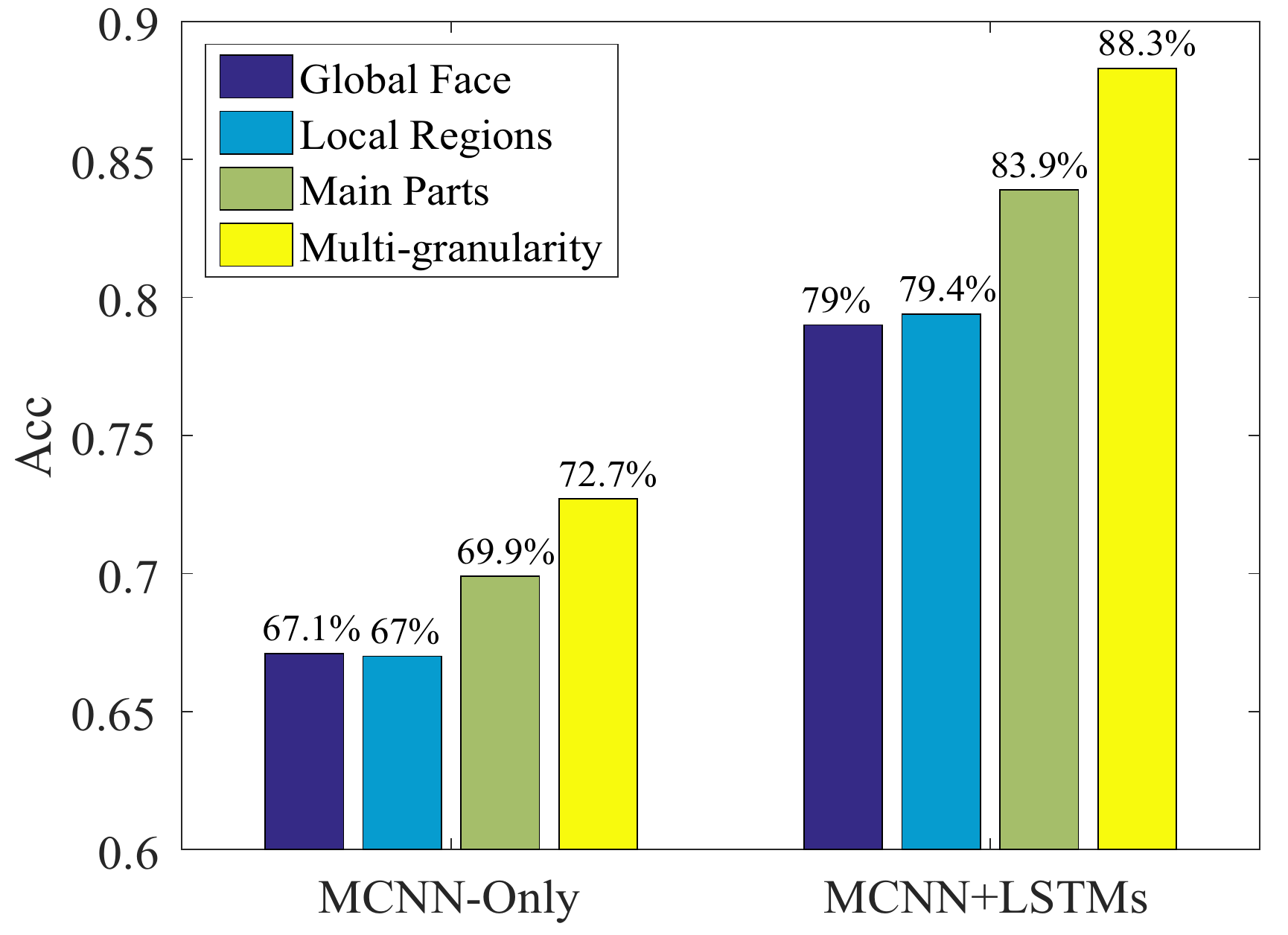}
	\caption{The comparison of accuracies achieved via MCNN only and MCNN+LSTMs on the testset of FI-DDD dataset. Different granularities are still presented.}
	\label{fig_cnn_rnn_multi}
\end{figure}

{\bf{MCNN-Only $vs$ MCNN+LSTMs:}}
The experiments are carried on four different granularities to research the effects of the multi-granularity and LSTMs. Fig.\ref{fig_cnn_rnn_multi} shows the accuracy of MCNN only and MCNN+LSTMs for detecting videos on testset under different granularities. MCNN-Only method obtains $72.7\%$ accuracy, while the accuracy achieved by MCNN+LSTMs is $15.6\%$ more than that by MCNN-Only. The reason is that the LSTMs have ability to mine the clues in temporal dimension which is significant for recognizing lots of ambiguous states, such as closing eyes and blinking. Comparing the accuracies of different granularities, we discover that the well-aligned multi-granularity facial patches still achieve the best performance. The accuracy of the main parts ranks the second, which means the granularity of main parts certainly plays the most important role in improving the effectiveness compared to the other two granularities.

\subsection{Comparisons with The Previous Methods}
We evaluate the whole method on the evaluation set and compare with the previous methods \cite{YuPLJ16,ParkPKY16,ShihH16} achieved on the same dataset. Due to the long-term memory characteristics on NTHU-DDD dataset, the max memory length is set to 120 frames and other parameters keep the same as the above experiments. Especially for night scenarios, we retrain a model with the night data of NTHU-DDD to detect driver drowsiness on near-infrared videos.  

{\bf{Accuracy:}}
Table \ref{tab_performance} presents the comparison of our method and the previous work \cite{YuPLJ16,ParkPKY16,ShihH16} and the proposed method achieves $90.05\%$ accuracy, as the state-of-the-art method of driver drowsiness detection.

{\bf{Speed:}}
We measure time consumption of all modules of our proposed method. From Table \ref{tab_time_consumption}, CNN costs the most time and the approach achieves about 3 fps on CPU platform. While on GPU platform, the proposed method can achieve 37 fps and satisfy the real-time performance requirements.
\begin{table}[htb]
	\centering
	\caption{Time consumption of each modules of the proposed method. Others include reading, writing and some converting operations.}
	\begin{tabular}{c|p{0.4cm}<{\centering}p{0.5cm}<{\centering}p{0.7cm}<{\centering}p{0.6cm}<{\centering}p{0.6cm}<{\centering}}
		\hline
		\diagbox{Platform}{Time(ms)}{Module} 	& \tabincell{c}{Mg}	& CNN		& LSTMs		& Others 	& Total \\
		\hline
		\rowcolor{tableColor}
		CPU(E5)+GPU(M40) 						& 11.1				& 10.7		& 0.6		& 4.5 		&   26.9 \\
		CPU(I7)									& 5.6 				& 302.3		& 0.6		& 3.2 		&   311.7 \\	
		\hline
	\end{tabular}
	\label{tab_time_consumption}
\end{table}

\section{Conclusion}
We propose an effective and efficient Long-term Multi-granularity Deep Framework to detect driver drowsiness on videos. 
Well alignment of facial patches ensures the effectiveness of representations under large pose variation, and multi-granularity patches efficiently concentrate on the most significant regions. 
Multi-granularity Convolutional Neural Network (MCNN) can effectively learn both detailed appearance information of the main parts and local to global spatial constraints. 
The deep Long Short Term Memory (LSTM) network works well on learning the temporal dependencies of spatial representations for driver drowsiness detection.

Moreover, we build a dataset named Forward Instant Driver Drowsiness Detection with higher precision of drowsy locations in temporal dimension. 
The dataset performs well in training model parameters and analyzing effects of several factors. 
Finally, we evaluate our method on the evaluation set of NTHU-DDD dataset and achieve $90.05\%$ accuracy and about 37 fps speed as the state-of-the-art method on driver drowsiness detection.

\bibliographystyle{IEEEtran}
\bibliography{egbib}

\end{document}